\documentclass[pmlr]{jmlr}

\usepackage{longtable}

\usepackage{booktabs}
\usepackage[load-configurations=version-1]{siunitx} 

\usepackage{cleveref}
\crefformat{section}{\S#2#1#3} 
\crefformat{subsection}{\S#2#1#3}
\crefformat{subsubsection}{\S#2#1#3}

\theorembodyfont{\upshape}
\theoremheaderfont{\scshape}
\theorempostheader{:}
\theoremsep{\newline}

\usepackage{tikz}
\usetikzlibrary{automata, positioning, arrows}
\tikzset{->, node distance=3cm}

\jmlryear{2023}
\jmlrworkshop{Proceedings of the 16th ICGI}

\title[Learning Transductions and Alignments with RNN Seq2seq Models]{Learning Transductions and Alignments \\ with RNN Seq2seq Models}

  \author{\Name{Zhengxiang Wang} \Email{zhengxiang.wang@stonybrook.edu} \\ 
          \addr Department of Linguistics \\ Institute for Advanced Computational Science \\ Stony Brook University}

\begin{document}

\maketitle


\begin{abstract}
The paper studies the capabilities of Recurrent-Neural-Network sequence to sequence (RNN seq2seq) models in learning four transduction tasks: identity, reversal, total reduplication, and quadratic copying. These transductions are traditionally well studied under finite state transducers and attributed with increasing complexity. We find that RNN seq2seq models are only able to approximate a mapping that fits the training or in-distribution data, instead of learning the
underlying functions. Although attention makes learning more efficient and robust, it does not overcome the out-of-distribution generalization limitation. We establish a novel complexity hierarchy for learning the four tasks for attention-less RNN seq2seq models, which may be understood in terms of the complexity hierarchy of formal languages, instead of string transductions. RNN variants also play a role in the results. In particular, we show that Simple RNN seq2seq models cannot count the input length.  \\
\end{abstract}

\begin{keywords}
RNNs, sequence to sequence, alignments, string transductions, attention, generalization abilities, complexity hierarchy, formal language theory 
\end{keywords}

\section{Introduction}
\label{sec:intro}

Given the black-box nature of neural networks, learning formal languages has emerged as ideal proxy tasks for evaluating the expressive power and generalization capacity of neural networks \citep{NIPS2015_b9d487a3, jeff-dnn, bhattamishra-etal-2020-ability, nn_chomsky, vanderpoel2023mlregtest}. Unlike real-world learning tasks, formal language tasks as well as their underlying functions are typically well-understood and known in advance. This makes possible a more flexible and complete control over data and, as a result, more fine-grained analyses of results obtained in experiments. Moreover, the rich tradition of studying formal languages offers critical insights into interpreting the learning results of neural networks, such as, from automata-theoretic perspectives \citep{peng-etal-2018-rational, rnn_automata, pmlr-v93-ayache19a, nn_chomsky}.

The current paper examines the learning capabilities of Recurrent-Neural-Network sequence to sequence (RNN seq2seq) models in the context of string transductions. It focuses on the abilities of various configurations of RNN seq2seq models to learn a series of four transduction tasks of increasing complexity and generalize to unseen in-distribution and out-of-distribution examples. The four tasks include identity, reversal, total reduplication, and quadratic copying (see \cref{sec:tasks} for a review). Previous works on RNN seq2seq models have only investigated identity \citep{NIPS2015_b9d487a3}, reversal \citep{NIPS2015_b9d487a3, rnn_automata}, and total reduplication \citep{nelson-etal-2020-probing}, but not quadratic copying. To the best of our knowledge, this paper is the very first study that analyzes the learning capabilities of three major RNN seq2seq model variants, with and without attention, on these four transduction tasks altogether. Both in-distribution and out-of-distribution generalization abilities are studied to better understand models' generalization capacity, which is central to a more rigorous and interpretable science of machine learning and its reliable application in real world \citep{ml_science}. The concept of distribution is closely related to the input sequence lengths \citep{nn_chomsky}, as detailed in \cref{sec:distribution}.

Through controlled and reproducible experiments, we present the very first comparative and comprehensive results of RNN seq2seq models with various configurations learning the four transduction tasks. These results allow us to draw a wide range of conclusions about the learning capabilities of these models and some contributing factors, such as attention, task complexity, and RNN variants. This study showcases the fruitfulness of probing neural networks with formal string transductions, a nascent approach \citep{nn_chomsky}.

The paper proceeds as follows. We review essential technical preliminaries about RNN seq2seq models in \cref{sec:preliminaries} and introduce the four learning tasks in \cref{sec:tasks}. Experimental setups, including data, model training details, and evaluation methods, are described in \cref{sec:setups}. We present the results in \cref{sec:results}, which are summarized and further discussed in \cref{sec: discussion}.

The source code, data, model training logs, trained models, and experimental results are open-sourced at \url{https://github.com/jaaack-wang/rnn-seq2seq-learning}.

\section{\label{sec:preliminaries}Preliminaries}

\subsection{RNNs}

RNNs represent a neural network architecture that utilizes repeated application of a recurrent unit to process a variable-length sequence $\mathbf{x} 
 = (x_1,...,x_T)$. At each time step $t$, the recurrent unit computes a vector $\mathbf{h}_t \in \mathbb{R}^{D \times 1}$ by taking as inputs the embedding $\mathbf{e}_{t}$ of the current input symbol $x_t$ (via an embedding layer $\mathbf{E}$) and the previous hidden state $\mathbf{h}_{t-1}$

\begin{equation}\label{eq:rrn_hidden}
\mathbf{h}_t = f(\mathbf{h}_{t-1}, \mathbf{e}_{t})
\end{equation}

\noindent
where $f(\cdot)$ is a non-linear state transition function and varies among different variants of RNNs. The hidden state is commonly initialized as a zero vector. A non-initial hidden state $\mathbf{h}_t$ may be passed to an output layer to compute the probability distribution of the next symbol $\hat{x}_{t+1}$ over an output alphabet $\Gamma$ of size $N$, using \emph{softmax} 

\begin{equation}\label{eq:rrn_output}
p(\hat{x}_{t+1, i}=1 \; | \; x_{t},...,x_{1}) = \frac{exp(\mathbf{w}_{i}\mathbf{h}_{t})}{\sum_{i'=1}^{N} exp(\mathbf{w}_{i'}\mathbf{h}_{t})}
\end{equation}

\noindent 
where $\hat{x}_{t+1, i}=1$ denotes $\hat{x}_{t+1}$ being the $i_{th}$ symbol in $\Gamma$ using one-hot encoding and $\mathbf{w}_i \in \mathbb{R}^{1 \times D}$ is a weight vector associated with that symbol. For the purpose of sequence generation, the embedding $\hat{\mathbf{e}}_{t+1}$ for $\hat{x}_{t+1}$ along with $\mathbf{h}_t$ can be passed as inputs to the recurrent unit to compute the subsequent hidden states and output symbols via the iterative or auto-regressive application of Eq.(\ref{eq:rrn_hidden}) and Eq.(\ref{eq:rrn_output}).

This study uses three most common variants \citep{yoavpremier}: Simple RNN (SRNN, \citealp{srnn}), Long Short-term Memory (LSTM, \citealp{lstm}), and Gated Recurrent Units (GRU, \citealp{cho-etal-2014-learning}). We choose these three variants because they are among most well recognized or popular RNNs to date. The main difference among these three types of RNNs lie in the construction of the recurrent unit, where LSTM and GRU come with additional gating mechanisms to control information flow across time steps, and LSTM has a cell state besides the hidden state. The mathematical details of the state transition functions for each type of RNN are provided in Appendix \hyperref[app:rnnfs]{A}. For simplicity and interpretability, all RNNs in this study are single-layered and unidirectional.

\subsection{RNN seq2seq models\label{sec:rnn_seq2seq}}

A RNN seq2seq model is an encoder-decoder structure where both the encoder and decoder are RNNs \citep{seq2seq, cho-etal-2014-learning}. Given a pair of variable-length sequences $\mathbf{x} = (x_1,...,x_T)$ and $\mathbf{y} = (y_1,...,y_{T'})$, the encoder consumes the input sequence $\mathbf{x}$ sequentially until the final hidden state $\mathbf{h}_T^{enc}$ is produced. The decoder takes as initial inputs $\mathbf{h}_T^{enc}$ and a preset start symbol $<\!s\!>$ and is trained to auto-regressively generate an output sequence $\hat{\mathbf{y}} = (\hat{y}_1,...,\hat{y}_{T'})$\footnote{This is only for training where $|\hat{\mathbf{y}}|$ is typically set equal to $|\mathbf{y}|$ in practice to achieve parallel computation. At inference, $|\hat{\mathbf{y}}|$ can differ from $|\mathbf{y}|$, depending on when the preset end symbol is generated.} to approximate $\mathbf{y}$ as much as possible. A preset end symbol $<\!/s\!>$ is also used to signal the termination of generation. Both the start and end symbols are appended to $\mathbf{x}$ and $\mathbf{y}$ in our experiments.  

In this study, we train three types of RNN seq2seq models, i.e., SRNN seq2seq, GRU seq2seq, and LSTM seq2seq, where the encoders and decoders are RNNs of same variant and with same hidden size. All the models are trained end-to-end by minimizing the cross-entropy loss between $\hat{\mathbf{y}}$ and $\mathbf{y}$ through mini-batch gradient descent.

\subsection{\label{sec:attention}Attention}

Attention\footnote{Here we only consider the so-called ``global attention" where the encoder's hidden states are all accessible.} is a mechanism that allows the decoder in a seq2seq model to access information from all hidden states $\mathbf{H}^{enc} \in \mathbb{R}^{D \times T}$ of the encoder. It is first proposed to improve the performance of neural machine translation \citep{jointly_align, luong-etal-2015-effective} and has later on been found to be a critical component of the Transformer architecture \citep{tranformer}. Attention has been hypothesized as external memory resources \citep{neuralTuringMachine} or a ``weighted skip connection” \citep{britz-etal-2017-massive} to account for the success of seq2seq models augmented with attention.

Formally, attention typically works as follows. At each decoding time step $t$, an attentional weight vector $\mathbf{a}_{t} \in \mathbb{R}^{T \times 1}$ can be computed by 

\begin{equation}\label{eq:attn}
\mathbf{a}_{t,i} = \frac{exp(score(\mathbf{h}_{t}^{dec}, \mathbf{h}_{i}^{enc}))}{\sum_{i'=1}^{T} exp(score(\mathbf{h}_{t}^{dec}, \mathbf{h}_{i'}^{enc}))}
\end{equation}

\noindent 
where $\mathbf{a}_{t,i}$ is a scalar weight that corresponds to the $i_{th}$ hidden state $\mathbf{h}_{i}^{enc}$ of the encoder and $score$ a function that measures how well $\mathbf{h}_{t}^{dec}$ aligns with $\mathbf{h}_{i}^{enc}$ for $i \in \{1,...,T\}$. A context vector $\mathbf{c}_{t} \in \mathbb{R}^{D \times 1}$ can be computed by weighing $\mathbf{H}^{enc}$ with $\mathbf{a}_{t}$ through matrix multiplication, then concatenated with the embedding for $\hat{{y}}_{t}$, and together consumed by the decoder to generate an output. There are many variants of the $score$ functions as in Eq.(\ref{eq:attn}) \citep{luong-etal-2015-effective, tranformer}. This study uses a simple one as follows

\begin{equation}\label{eq:attn_score}
score(\mathbf{h}_{t}^{dec}, \mathbf{h}_{i}^{enc}) = \mathbf{v}_{a} \; tanh(\mathbf{W}_{a}[\mathbf{h}_{t}^{dec};\mathbf{h}_{i}^{enc}])
\end{equation}

\noindent 
where $\mathbf{W}_{a} \in \mathbb{R}^{D\times 2D}$ and $\mathbf{v}_{a} \in \mathbb{R}^{1 \times D}$ are learnt weights to reduce the concatenated hidden states $[\mathbf{h}_{t}^{dec};\mathbf{h}_{i}^{enc}] \in \mathbb{R}^{2D \times 1}$ to an alignment score, and $tanh$ is a hyperbolic tangent function.

\subsection{RNNs versus RNN seq2seq models\label{sec:RNN_vs_RNN_seq2seq}}

RNNs are often compared to finite state automata \citep{Giles1992, Siegelmann1996, Visser2000, chen-etal-2018-recurrent, rnn_automata} because RNNs consume an input sequence to produce a membership decision. To model transductions, RNNs write an output symbol right after reading an input symbol, just like finite state transducers (FSTs). RNN seq2seq models, however, consume all the input symbols before producing any output symbols. The two distinct ways of processing the inputs and generating the outputs form two different classes of neural networks. Consequently, previous results based on RNNs should not be confused with the current results based on RNN seq2seq models. 

That said, both RNNs and RNN seq2seq models are sequential models. When trained with backpropagation and gradient descent, the long-term dependency learning issue identified in RNNs \citep{bengio_long_dep_issue} is also an issue for attention-less RNN seq2seq models without the skip connections \citep{britz-etal-2017-massive} introduced by attention. This is because the backpropagated gradients may either decay or grow exponentially, resulting in the notorious problems of vanishing or exploding gradients \citep{pmlr-v28-pascanu13, Chandar2019}. When either happens, weight optimization becomes impossible, which prevents learning.

\section{\label{sec:tasks}Learning tasks}

\subsection{Task description and FST characterizations\label{sec:task_desc}}

We are interested in the following four learning tasks, representable by four deterministic string-to-string functions with an input alphabet $\Sigma$ and an output alphabet $\Gamma$: (A) identity; (B) reversal; (C) total reduplication; (D) quadratic copying. For a given string $w \in \Sigma^*$, $f_{A}(w) = w$, $f_{B}(w) = w^R$, $f_{C}(w) = ww$, and $f_{D}(w) = w^{|w|}$, where $w^R$ denotes the reverse of \emph{w}. For example, if $w = abc$, then $f_{A}(abc) = abc$, $f_{B}(abc) = cba$, $f_{C}(abc) = abcabc$, and $f_{D}(abc) = (abc)^3 = abcabcabc$. Let $|w| = n$ and $|f_{D}(w)| = n^2$. Hence the name of quadratic copying or squaring function for $f_{D}$. For all these functions, $\Sigma = \Gamma$.

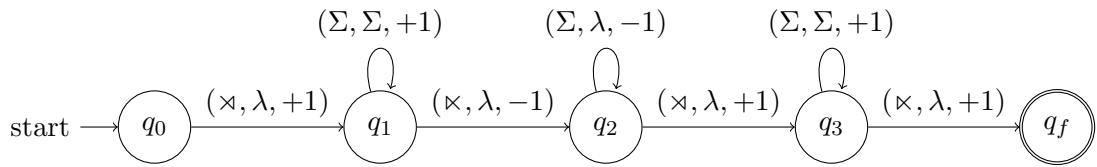
\begin{figure}[ht]
    \centering
    \begin{tikzpicture}
        \node[state, initial] (q1) {$q_0$};
        \node[state, right of=q1] (q2) {$q_1$};
        \node[state, right of=q2] (q3) {$q_2$};
        \node[state, right of=q3] (q4) {$q_3$};
        \node[state, accepting, right of=q4] (q5) {$q_f$};

    \draw 
        (q1) edge[above] node{$(\rtimes, \lambda, +1)$} (q2)
        (q2) edge[loop above, above] node{$(\Sigma, \Sigma, +1)$} (q2)
        (q2) edge[above] node{$(\ltimes, \lambda, -1)$} (q3)
        (q3) edge[loop above, above] node{$(\Sigma, \lambda, -1)$} (q3)
        (q3) edge[above] node{$(\rtimes, \lambda, +1)$} (q4)
        (q4) edge[loop above, above] node{$(\Sigma, \Sigma, +1)$} (q4)
        (q4) edge[above] node{$(\ltimes, \lambda, +1)$} (q5);
    \end{tikzpicture}
    \caption{An example 2-way FST to model total reduplication $f_{C}: w \rightarrow ww$, with $w$ padded into $\ltimes w \ltimes$ as the input. $\lambda$: empty string; $+1$: move right; $-1$: move left.}
    \label{fig:ex_2way_fst}
\end{figure}

Traditionally, the four tasks are modelled with FSTs \citep{transducers_logisc_algebra, bojanczyk_et_al_2019, Dolatian_Heinz_2020, regular_polyregular_2023}. More concretely, $f_{A}$ can be easily modelled by a 1-way FST where each input symbol is simply mapped to itself, whereas a 2-way FST that can read input sequences back and forth is used for modelling $f_{B}$ and $f_{C}$, such as the one in Fig~\ref{fig:ex_2way_fst}. To model $f_{D}$, a 2-way FST enhanced with the capability of counting the length of the input string is needed. As these four tasks require FSTs of increasing expressive capacity, they are characterized accordingly \citep{regular_polyregular_2023}, with $f_{A}$ being a rational function, $f_{B}$ and $f_{C}$ regular functions, and $f_{D}$ a polyregular function \citep{bojanczyk_et_al_2019}. Under the FST-theoretic characterizations, $f_{D} > f_{C} > f_{B} > f_{A}$, where $>$ is a “more complex than” relation. Although $f_{B}$ and $f_{C}$ are both regular functions, a 2-way FST has to scan any input sequence three times (forward-backward-forward) to model $f_{C}$, instead of two times (forward-backward) to model $f_{B}$.

\subsection{In-distribution and out-of-distribution\label{sec:distribution}}

Given the deterministic nature of the four functions above, we define the concept of in-distribution and out-of-distribution in terms of the input sequences. For a model trained on input sequences of lengths $\mathcal{L}$, in-distribution input sequences are those whose lengths $\mathcal{L'} \subseteq \mathcal{L}$. In the context of this study, input sequences are out-of-distribution if $\mathcal{L'} \cap \mathcal{L} = \O$. For example, in the experiments described in the next section, strings with lengths between 6 and 15 (inclusive) constituted in-distribution sequences, and other lengths are out-of-distribution.

Distinguishing in-distribution and out-of-distribution input sequences allows us to examine a trained model's ability to generalize to examples that are independent and identically distributed and those that are beyond, in relation to the distribution of the training examples. Furthermore, a trained model's out-of-distribution generalization ability reveals whether the model learns the underlying function or approximates the in-distribution data.

\subsection{Complexity hypothesis\label{sec:complexity_hypothesis}}

\begin{figure}[!htb]
\centering
  \includegraphics[width=1\columnwidth]{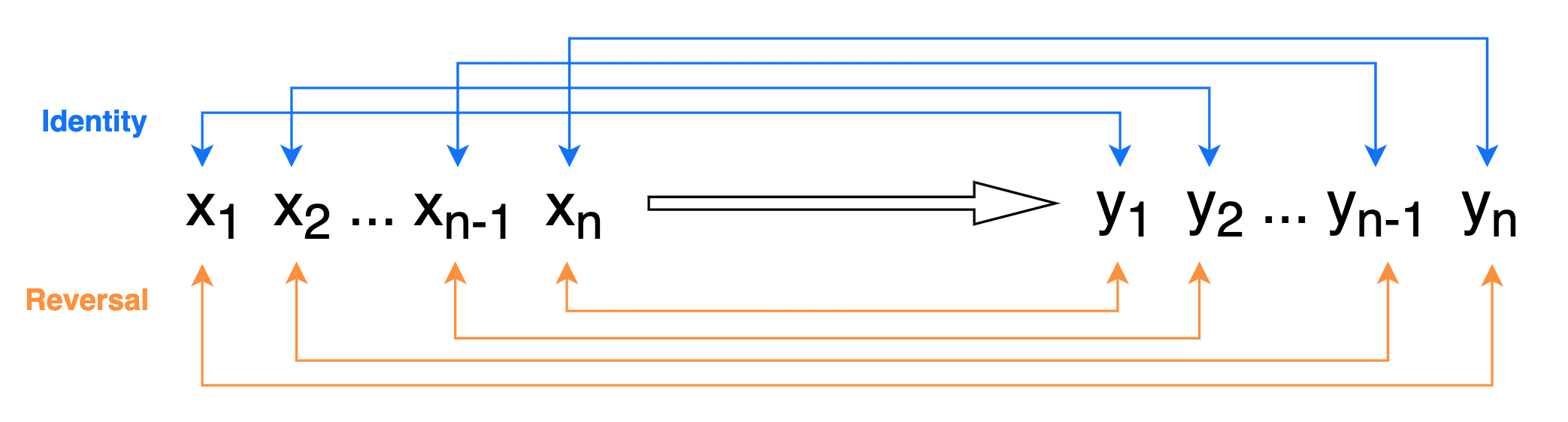}
\caption{The conjectured mechanism for RNN seq2seq models learning identity and reversal. The multiple crossings on the top relate to identity; the multiple nested crossings at the bottom relate to reversal.}
\label{fig:learning_alignments}
\end{figure}

As discussed in \cref{sec:RNN_vs_RNN_seq2seq}, RNN seq2seq models take an encoder-decoder structure, where the decoder only writes after the encoder read all the input symbols, unlike the read-and-write operation seen in FSTs or RNNs. Therefore, for a RNN seq2seq model to learn these tasks, the decoder must recall all the input symbols from the information passed from the encoder and select the output symbols in correct alignments with the input symbols. In other words, the four tasks require RNN seq2seq models to learn varying input-target alignments or dependencies. Fig~\ref{fig:learning_alignments} illustrates the conjectured mechanism for learning identity and reversal. Total reduplication and quadratic copying can be learnt in a similar process, as the outputs of these two functions can be seen as the concatenation of multiple identity functions applied in a sequence \citep{regular_polyregular_2023}. To learn quadratic copying, the model should additionally be able to count the length of the input sequence. 

Accordingly, we propose the following task complexity hierarchy for attention-less RNN seq2seq models: quadratic copying ($f_{D}$) $>$ total reduplication ($f_{C}$) $>$ identity ($f_{A}$) $>$ reversal ($f_{B}$). Unlike the hierarchy established in \cref{sec:task_desc} under FSTs where $f_{B} > f_{A}$, this hierarchy claims that $f_{A} > f_{B}$. As can be seen in Fig~\ref{fig:learning_alignments}, given the same input sequence of length $n$, the input-target dependencies for $f_{A}$ are of constant distance, i.e., $n$. However, there are initially $\frac{n}{2}$ or so shorter dependencies for $f_{B}$, which leads to easier weight optimization with the iterative application of backpropogation and gradient descent \citep{seq2seq}. When the long-term dependency learning issue discussed in \cref{sec:RNN_vs_RNN_seq2seq} becomes an issue, this results in less complexity in learning $f_{B}$ than $f_{A}$. Similarly, since $f_{C}$ and $f_{D}$ introduces more and longer dependencies than $f_{A}$, they are necessarily harder. Also, with longer target sequences for given input sequences requires more expressive power to distinguish the right output sequences among exponentially growing possible candidates. For this reason, $f_{D}$ is the most complex to learn since its target sequence length grows quadratically as a function of the input length, instead of linearly as for $f_{A}$ and $f_{B}$. 

The complexity hypothesis proposed above may be understood in terms of the complexity hierarchy of formal languages, if we combine the input and target sequences together and take the transduction tasks as language recognition tasks. For example, if we replace the arrow in Fig~\ref{fig:learning_alignments} with \# and concatenate the input and target sequences, we have a copy language $w\#w$ for $f_{A}$ and a palindrome language $w\#w^R$ for $f_{B}$. According to the Chomsky’s hierarchy \citep{chomsky}, these two languages belong to context sensitive language (CSL) and context free language (CFL), respectively, where CSL is more complex than CFL. Similarly, the two respective languages corresponding to $f_{C}$ and $f_{D}$ are $w\#ww$ and $w\#w^{|w|}$, where $w\#ww$ as well as $w\#w$ is considered a mildly CSL \citep{joshi_1985} whereas $w\#w^{|w|}$ a strict CSL. Obviously, under the complexity hierarchy of formal languages, the languages associated with the four functions also imply that $f_{D} > f_{C} > f_{A} > f_{B}$. 

For attentional models, however, we hypothesize that $f_{D} > f_{C} > f_{A} \geq f_{B}$, where $\geq$ means ``no less complex than". This is because attention allows the decoder to access all the encoder's hidden states at any decoding time step, potentially alleviating the long-term dependency learning issue identified in the RNNs without attention.

\section{\label{sec:setups}Experimental setups}

To ensure a fair evaluation of the learnability of the four tasks by RNN seq2seq models in a finite setting, we equipped all models with a decently large parameter size such that the lack of sufficient capacity to fit the training data is less of a bottleneck. For the same reason, we also utilized various training techniques to improve the success rate of convergence for all the models. To make the results comparable across models of varying configurations and across different tasks, the input sequences and the training and evaluation conditions were deliberately set identical for every model trained and evaluated.

\subsection{\label{sec:data}Data}

To simulate a more realistic learning setting, we set $\Sigma$ and $\Gamma$ both to be the 26 lowercase English letters. For every $l \in \{1, ..., 30\}$, we randomly sampled from $\Sigma^{l}$ the same number of strings as the input sequences, where duplicates are only allowed for sequences of lengths 1 and 2. The target sequences were obtained by applying the four deterministic functions that represent the tasks. In-distribution strings are those of input lengths 6-15, available in the train, dev (development), and test sets. Out-of-distribution strings are those of input lengths 1-5 and 16-30, available only in the gen (generalization) set. 

For the train/dev sets, there are 1,000 input-target pairs per input length. For test/gen sets, the number is 5,000 per input length. The test and gen sets were made five times larger than the train/dev sets for the sake of more reliable evaluations. The four datasets are mutually disjoint. More details about the data can be found in Appendix \hyperref[app:data]{B.1}.

\subsection{\label{sec:training}Training details}

Models were constructed in PyTorch \citep{pytorch} and run on Google Colaboratory. The single-layered RNN encoder and decoder in each model are both of hidden size 512 and contain an embedding layer of embedding size 128. We initialized all trainable weights using Xavier initialization \citep{pmlr-v9-glorot10a} to reduce the vanishing gradient problem. We applied Adam optimizer \citep{adam} with 5e-4 learning rate and 1e-5 L2 weight decay rate. We normalized gradients with maximum norm clipped at 1 \citep{pmlr-v28-pascanu13} to alleviate or circumvent the exploding gradient problem \citep{DBLP:conf/iclr/ZhangHSJ20}. Details about the model parameter configuration and size are provided in Appendix \hyperref[app:model_size]{B.2}. 

To speed up convergence at the training time, we employed a technique called teacher forcing \citep{teacher-forcing} to permit the decoders to access the real next symbols from the target sequences, instead of using the predicted ones as inputs. All models were trained up to 500 epochs with the train/dev sets performances evaluated every 10 epochs. The batch size is 1,000 and every batch only contained input sequences of same lengths to avoid padding, which changes the input-target mapping. Training only stopped if one of the following conditions was met: (1) models trained through all the epochs; (2) the full-sequence accuracy (see \cref{sec:eval_methods}) in the dev set reaches exactly 100\% with double precision; (3) the full-sequence accuracy in the train set and dev set exceeds 99.99\% and 99.50\% simultaneously. Every model was trained and evaluated for 3 runs. In each run, the model with the highest full-sequence accuracy on the dev set was saved and deployed to the test and gen sets. Appendix \hyperref[app:training_notes]{B.3} provides some important training notes.

\subsection{\label{sec:eval_methods}Evaluation methods}

We used the following three metrics to evaluate how well RNN seq2seq models learn the input-target alignments for the four tasks: full-sequence accuracy, first $n$-symbol accuracy, and overlap rate. All these metrics are measured from the initial symbol to the end symbol $<\!/s\!>$ of the target sequences against the corresponding output sequences generated to the same lengths (for efficient training within batches). Full-sequence accuracy measures the percentage of the target sequences being correctly generated as a whole, whereas first $n$-symbol accuracy measures the average proportion of the first $n$ symbols being correctly generated for the target sequences. Overlap rate is the average pairwise overlapping ratio of the output sequences to the target sequences.

These three metrics provide well-rounded measurements of alignments between two sequences. When a more restrictive metric shows a low score and thus becomes less discriminative, there can be a more fine-grained alternative, because neural networks do not learn the input-target alignments strictly from the very left incrementally to the very right, due to the random initialization of model parameters. In this study, we used the full-sequence accuracy as the main metric and reported the last two metrics when necessary.

\begin{table*}[ht]
\centering
\small
  {\caption{\label{tab:main_res_full_seq_accu}Aggregate full-sequence accuracy (\%) across the four learning tasks for models with various configurations. Best results are in \textbf{bold} for the test and gen sets.}}
  \vspace{10pt}
  {\begin{tabular}{llrrrrrrrr}
  \toprule
  \multicolumn{2}{c}{} & & \multicolumn{3}{c}{\bfseries Attentional} & & \multicolumn{3}{c}{\bfseries Attention-less} \\ \hline
\bfseries Task & \bfseries Dataset & & \bfseries SRNN & \bfseries GRU & \bfseries LSTM & & \bfseries SRNN & \bfseries GRU & \bfseries LSTM \\ \hline

& Train & & 100.00 & 100.00 & 100.00 & & 69.74 & 98.26 & 100.00 \\ 
Identity & Test & & 99.97 & \bfseries 100.00 & \bfseries 100.00 & & 42.82 & 70.46 & \bfseries 77.57 \\ 
& Gen & & 25.52 & \textbf{37.41} & 36.37 & & 0.00 & \textbf{10.41} & 10.01 \\ \hline

& Train & & 100.00 & 100.00 & 100.00 & & 100.00 & 100.00 & 100.00 \\ 
Rev & Test & & \bfseries 99.98 & 99.87 & 99.88 & & \bfseries 99.55 & 88.46 & 92.85 \\ 
& Gen & & \textbf{40.14} & 23.54 & 25.79 & & \textbf{23.89} & 19.72 & 12.42 \\ \hline

& Train & & 100.00 & 100.00 & 99.99 & & 15.22 & 90.57 & 93.51 \\ 
Total Red & Test & & 99.71 & \bfseries 99.77 & 99.64 & & 5.60 & 50.76 & \bfseries 55.17 \\ 
& Gen & &  \textbf{42.34} & 23.23 & 20.31 & & 0.00 & 4.39 & \textbf{6.18} \\ \hline

& Train & & 2.43 & 79.84 & 82.73 && 1.62 & 49.29 & 67.29 \\ 
Quad Copy  & Test & & 1.99 & 67.75 & \bfseries 73.89 && 0.61 & 27.76 & \bfseries 38.03 \\ 
& Gen & & 1.36 & \bfseries 8.20 & 6.07 && 0.00 & \bfseries 0.85 & 0.18 \\ \hline

& Train & & 75.61 & 94.96 & 95.68 && 46.65 & 84.53 & 90.19 \\ 
Average & Test & & 75.41 & 91.85 & \bfseries 93.35 && 37.15 & 59.36 & \bfseries 65.91 \\ 
& Gen & & \bfseries 27.34 & 23.10 & 22.13 && 5.97 & \bfseries 8.85 & 7.20 \\ 
  \bottomrule
  \end{tabular}}
\end{table*}

\begin{figure*}[ht]
\centering
  \includegraphics[width=1\columnwidth]{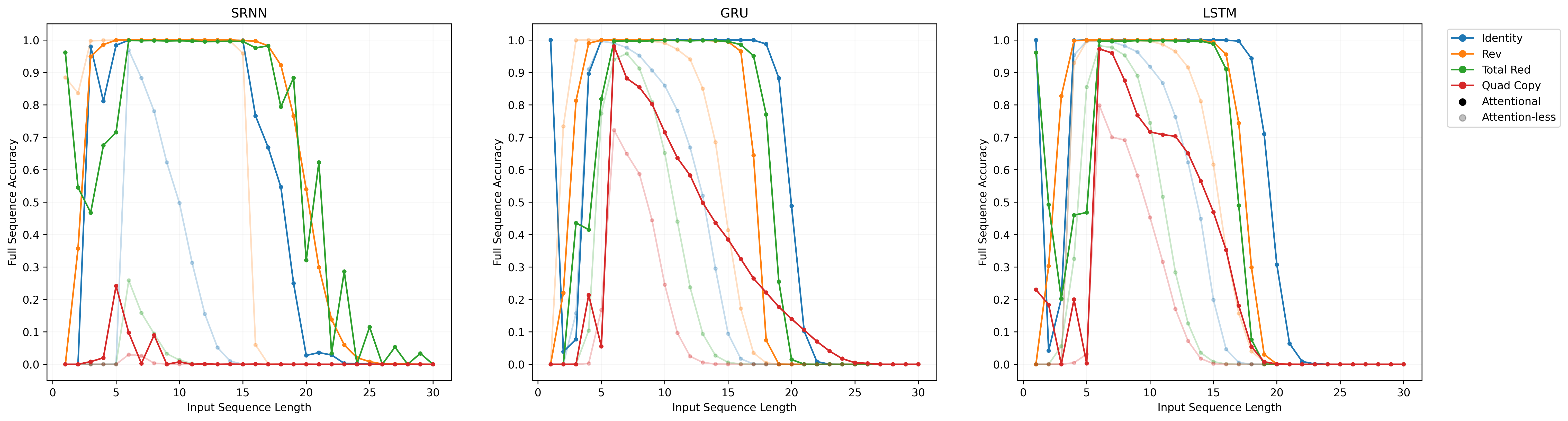}
\caption{Test/gen set full-sequence accuracy per input length across the four tasks for the three types of RNN seq2seq models, with and without attention. Test set length range: 6-15; gen set length range: 1-5 \& 16-30.}
\label{fig:full_seq_accu_per_len}
\end{figure*}

\section{\label{sec:results}Results}

The main results can be found in Table~\ref{tab:main_res_full_seq_accu} and Fig~\ref{fig:full_seq_accu_per_len}, which present full-sequence accuracy on aggregate and per-input-length levels, respectively. Related results measured in first $n$-symbol accuracy and overlap rate, are in Appendix \hyperref[app:results]{C} for references. Since the results across these three metrics share similar patterns, unless particularly mentioned, findings based on full-sequence accuracy remain valid in terms of the other two metrics. This section also reports results from follow-up experiments conducted under contrastive settings. 

Please note that, since we are only interested in the learning potentials of RNN seq2seq models, the results presented in the body text are the overall best results, instead of the average ones. The best results were selected out of the three runs based on the test and gen set performance in full-sequence accuracy, empirically weighted 40\% and 60\%, respectively. This is consistent with our training procedure, where only the models with best dev set performance are deployed and evaluated, and in line with \cite{NIPS2015_b9d487a3} and \cite{nn_chomsky} who also choose to report best results. Nevertheless, the observations based on the best results are still true for the average results (see Appendix \hyperref[app:avg_full_seq_accu]{C.1}).

\subsection{Generalization abilities\label{sec:gen_behaviors}}

As can be observed in Table~\ref{tab:main_res_full_seq_accu}, RNN seq2seq models, with or without attention, consistently achieve better aggregate full-sequence accuracy in the test sets than in the gen sets with a large average margin (i.e., at least 31\%) for all the tasks. Furthermore, even on the per-input-length level, depicted in Fig~\ref{fig:full_seq_accu_per_len}, the gen set full-sequence accuracy decreases as the difference between the unseen lengths and the nearest seen lengths increases most of the time. These strongly demonstrate that RNN seq2seq models, regardless of attention, are prone to learning a function that fits the training or in-distribution data, instead of the underlying data generation functions, if these functions are learnable at all. Their out-of-distribution generalization abilities are highly limited if longer unseen and untested input sequence lengths are also taken into account. Empirically, out-of-distribution generalization appears to be restricted to data that is more similar to the training data in lengths.

As a result of approximating in-distribution data, the gen set performance is also considerably less correlated with the train set performance, compared to the test set. On the one hand, the train-gen variance is consistently significantly larger than the train-test variance for every model in each run. On the other hand, by measuring the correlation in the standard derivation of full-sequence accuracy of each model across the three runs, we find that the Kendall/Spearman correlation scores for train-test and train-gen are 0.79/0.89 and -0.35/-0.52, respectively, confirming that the train and test set performance is more positively correlated. For instance, Table~\ref{tab:main_res_full_seq_accu_avg} shows that when attentional models fit the train set nearly perfectly ($\approx 100.00\%$) with a negligible performance variation in the test set ($\leq 0.12\%$), the variation in the gen set, however, ranges from 1.56\% to 10.03\%. 

Moreover, with the follow-up experiments in \cref{sec:attn_learn_res} and \cref{sec:task_complexity}, we find that the out-of-distribution generalization is not correlated with the model complexity either. That is to say, the gen set performance is not a monotonic function of the number of parameters of the trained models, even when they all can fit the train/test sets nearly perfectly. A possible explanation for this is that when the models have sufficient capacity to fit the train set, there are potentially many or infinite possible configurations of parameters, which, however, may lead to very different out-of-distribution performance. In this sense, the out-of-distribution performance is less interpretable than the in-distribution one.

The lack of correlation in performance between the train and gen set may be caused by the selection of the best models during training, which is informed by dev sets that consist of pure in-distribution data. The exact reasons, however, remain to be explored.

\subsection{Attentional versus attention-less models\label{sec:attn_learn_res}}

The main results show straightforwardly that attention helps models fit the train sets and generalize to the test/gen sets. Attentional models can always achieve much better aggregate full-sequence accuracy in both the train and test sets and have a much smaller train-test variance than the attention-less counterparts. Moreover, attentional models outperform the attention-less models in generalizing to the out-of-distribution examples. In other words, attentional RNN seq2seq models are stronger in-distribution learners with relatively better out-of-distribution generalization abilities, compared to the attention-less ones. 

Besides, attention significantly improves learning efficiency, as attentional models tended to converge faster and used less than 50\% of the epochs on average than the attention-less models (see Appendix \hyperref[app:training_notes]{B.3}). Furthermore, Fig~\ref{fig:full_seq_accu_per_len} shows that the test set performance of the attention-less models goes down nearly as a function of the input length, which indicates the need of greater sample complexity for training. To further contrast the learning efficiency between attentional and attention-less models, we conducted a follow-up experiment in total reduplication, which is harder than identity and reversal, but is more feasible than quadratic copying to test for computational reasons. In the experiment, others being same, the attentional models used 1/4 training examples and 1/4 hidden size (i.e., 128), and the attention-less models used 3 times more training data and 3 times more training epochs, compared to the respective original setups. The results in Table~\ref{tab:attn_efficiency} show that by using only 1/12 training examples, 1/9 parameter size (see Appendix \hyperref[app:model_size]{B.2}), and 1/3 training epochs, the attentional models still outperform the attention-less ones in total reduplication.

\begin{table*}[ht]
\centering
\small
  {\caption{\label{tab:attn_efficiency}Aggregate full-sequence accuracy (\%) for the follow-up experiment in learning total reduplication. Training details are described in text in \cref{sec:attn_learn_res}.}}
  \vspace{10pt}
  {\begin{tabular}{lrrrrrrrr}
  \toprule
  & \multicolumn{3}{c}{\bfseries Attentional} & & \multicolumn{3}{c}{\bfseries Attention-less} \\ \hline
  
  \bfseries Dataset & \bfseries SRNN & \bfseries GRU & \bfseries LSTM & & \bfseries SRNN & \bfseries GRU & \bfseries LSTM \\ \hline
  Train & 100.00 & 100.00 & 100.00 & & 94.99 & 100.00 & 100.00 \\ 
   Test & 99.20 & 99.53 & 99.58 & & 84.93 & 90.21 & 91.86 \\ 
   Gen & 35.20 & 14.07 & 19.37 & & 0.00 & 5.10 & 4.54 \\
  \bottomrule
  \end{tabular}}
\end{table*}

\subsection{Task complexity\label{sec:task_complexity}}

The complexity hypothesis formulated in \cref{sec:complexity_hypothesis} for the four learning tasks is borne out for attention-less models. Table~\ref{tab:main_res_full_seq_accu} and Fig~\ref{fig:full_seq_accu_per_len} show clear evidence that for each type of attention-less models, quadratic copying is more complex than total reduplication, which is more complex than identity, with reversal being the least complex, on both aggregate and per-input-length levels. Given the identical training and evaluation conditions for the four tasks indicated in \cref{sec:setups}, we argue that the complexity of learning these tasks is the major, if not only, attributable reason for the observed performance difference. 

\begin{figure}[!htb]
\centering
  \includegraphics[width=1\columnwidth]{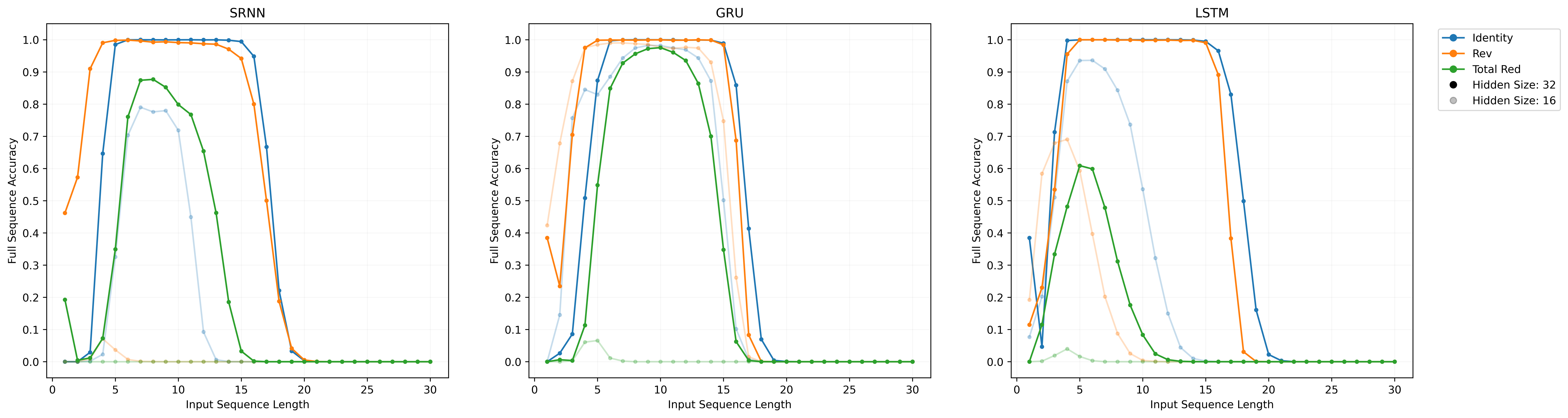}
\caption{Test/gen set full-sequence accuracy across the three tasks (except quadratic copying) for the three attentional RNN seq2seq models with two reduced hidden sizes: 16 and 32. The results for the hidden size 64 are not included because of the near 100\% and thus less informative train/test performance for all the tasks.}
\label{fig:full_seq_accu_per_len_follow_up}
\end{figure}

For attentional models, it is only clear that quadratic copying remains the most complex task to learn. For the rest tasks, however, the results do not distinguish their relative complexity informatively enough as full-sequence accuracy in the train and test sets all nears 100.00\%, due to apparent overparameterization. To verify the related complexity hierarchy proposed in \cref{sec:complexity_hypothesis}, we re-trained all the attentional models with three decreasing hidden sizes (i.e., 16, 32, 64, since \cref{sec:attn_learn_res} suggests that 128 is already too big) while keeping the rest original setups unchanged. The results in Fig~\ref{fig:full_seq_accu_per_len_follow_up} show that total reduplication is more complex than identity and reversal, since it requires more parameters or greater model complexity than the other two tasks. That said, Fig~\ref{fig:full_seq_accu_per_len_follow_up} also shows that when the hidden size is 16, the attentional SRNN and LSTM models appear to learn identity better than reversal, which contradicts with our proposal for unclear reasons. 


\subsection{RNN variants\label{sec:rnn_variants}}

Overall, when attention is not used, GRU and LSTM are expectedly and significantly more expressive than SRNN for all tasks other than reversal. This probably results from the additional built-in gating mechanisms that improve long-term memory \citep{lstm, gru_} and thus reduce the sample complexity to learn the input-target alignments. For reversal, we find that when the models fit the train set with (nearly) 100\% full-sequence accuracy, SRNN consistently shows better generalization abilities for reversal, both in-distribution and out-of-distribution, than GRU and LSTM, regardless of attention. In particular, when attention is used and hidden size $\geq 32$ (see the follow-up experiments in \cref{sec:attn_learn_res} and \cref{sec:task_complexity}), the worst performing SRNN always outperforms the best performing GRU and LSTM in the gen sets with other conditions being equal. For identity and total reduplication, however, such a consistent picture does not occur.

Previous research \citep{weiss-etal-2018-practical,  merrill-2019-sequential, nn_chomsky} has shown that LSTM can learn counting more effectively than SRNN and GRU. The results for quadratic copying largely support this claim, but also show that GRU may also learn counting in a similar way to LSTM in the seq2seq construction. To make sure that this is not due to the much smaller parameter size in the SRNN seq2seq models (see Appendix \hyperref[app:model_size]{B.2}), we re-trained them, both attentional and attention-less, with significantly increased model complexity. However, the best train set full-sequence accuracy among them is only 3.43\% (see Appendix \hyperref[app:attn_srnn_quad_copy]{D.2}), implying that the increased model complexity helps little.

Moreover, we notice that for quadratic copying, attentional SRNN models have unusually high and overall best first $n$-symbol accuracy and overlap rate (see Appendix \hyperref[app:other_main_results]{C.2}), despite rather low full-sequence accuracy. To confirm that SRNN does not count, we tested all the previously saved attentional models for quadratic copying on the mapping $w \rightarrow w^{40}$ (i.e., making 40 copies), which is not present in the data for quadratic copying. Not surprisingly, full-sequence accuracy is always 0.00\% for all the models. However, the best SRNN model achieves 96.20/80.81\% first $n$-symbol accuracy in the test/gen set, while the same figure for the GRU and LSTM models never exceeds 26.83/23.66\% (see Appendix \hyperref[app:attn_srnn_quad_copy]{D.2}). This means that the attentional SRNN models simply have learnt somehow periodically repeating the input sequences, which leads to the accidentally high first $n$-symbol accuracy and overlap rate for quadratic copying. Their inabilities to generate the end symbol $<\!s\!>$ at the right timing indicates their inabilities to count the input length. We encourage future studies to explore why they approximated a periodic repetition function instead.

\section{\label{sec: discussion}Discussion and Conclusion}

This study investigated how well the three major types of RNN seq2seq models, with and without attention, learn four transduction tasks that require learning varying alignments or dependencies between the input and target sequences. Through unified training/evaluation conditions and comprehensive experiments, we compared the learning results across tasks, different model configurations, and test/gen set performance etc., and highlighted factors that influence the learning capabilities and generalization capacity of RNN seq2seq models. Unlike previous research, the input alphabet $\Sigma$ for our experiments has 26 unique symbols, instead of binary, making our results more relevant to real-world tasks that concern, say, morpho-phonological transductions. The major findings are further discussed below. 

\textbf{Generalization abilities}. For our experiments, RNN seq2seq models, regardless of attention, tend to approximate the training or in-distribution data, instead of learning the underlying functions. This makes their out-of-distribution generalization abilities restricted, which, however, may not be surprising.\footnote{The definition of out-of-distribution sequences used in this paper may be too strict. In preliminary work, experiments on reversal and identity were conducted where the training data consisted of strings of lengths 11 to 30 and 41 to 60. Thus strings with lengths between 31 and 40 were technically out-of-distribution. However, the attentional models were able to generalize correctly on test items with unseen lengths within this gap, but not on test items with unseen lengths longer than 60.} For an auto-regressive model like RNN seq2seq models to learn these transductions with bounded precision or parameter size, the probability of the correct output decreases exponentially as a function of the output length $n$, i.e., $P(target) = (1-\varepsilon)^{n}$, where $\varepsilon$ is the expected error rate. As $n$ increases indefinitely, the probability of correct generation eventually becomes infinitesimal and indistinguishably small. It is true for both training and testing where strings of (unseen) longer lengths are generally more difficult to fit and generalize to, as evidenced by the results in~\cref{sec:results}.   

\textbf{Attention}. Attention greatly improves the learning efficiency, by reducing both model complexity and sample complexity, as well as the learning robustness, in terms of the generalization performance and overfitting problem. However, attention does not overcome the out-of-distribution generalization limitation, since it does not change the auto-regressive nature of the models. Nonetheless, the impressive learning efficiency accelerated by attention echoes its original motivation, i.e., ``learning to align" \citep{jointly_align}.

\textbf{Task complexity}. For the four tasks: identity ($f_A$), reversal ($f_B$), total reduplication ($f_C$), quadratic copying ($f_D$), we established the following task complexity hierarchy for attention-less RNN seq2seq models: $f_{D} > f_{C} > f_{A} > f_{B}$. This differs from the traditional FST-theoretic viewpoint, which sees $f_B$ strictly more complex than $f_{A}$. As discussed in \cref{sec:complexity_hypothesis}, $f_{B}$ is easier than $f_{A}$ for RNN seq2seq models because learning $f_{B}$ contains many initially shorter input-target dependencies, instead of constantly-distanced dependencies for $f_{A}$. This makes iteratively optimizing the model parameters easier with backpropogation and results in less complexity for $f_{B}$. Interestingly, we also show in \cref{sec:complexity_hypothesis} that this result may be understood under the complexity hierarchy of formal languages, if the four transduction tasks are re-framed as language recognition tasks. However, for attentional models, we find that $f_{D} > f_{C} > f_{B} > f_{A}$, which appears to align with the FST characterizations but contradicts with our expectation that $f_{A}$ and $f_{B}$ are comparably complex with attention. Constrained by time and resources, we leave the related investigation to future studies.

\textbf{RNN variants}. The effect of RNN variants on the seq2seq models is a complicated one and interacts with other factors, e.g., attention and the task to learn. When attention is
not used, GRU and LSTM are generally more expressive than SRNN. The only exception is reversal, which SRNN consistently outperforms GRU and LSTM in the test/gen set, regardless of attention. Please note that, this exception is only true when SRNN seq2seq models fit the related train set with (nearly) 100\% full-sequence accuracy. Moreover, for quadratic copying and regardless of attention, both GRU and LSTM can count the input length while SRNN cannot, which arguably is not a matter of model parameter size.

Since all the tasks in the study require the decoder to recall the entire input symbols at any decoding time step, an important step to test the generality of our conclusions is to experiment with tasks that do not. Inspired by one of the reviewers, we re-run the main experiments on ascending sorting ($f_E$), and descending sorting ($f_F$), which sort the input sequence in an ascending and descending lexicographic order, respectively. For example, for any string $w \in \{abc, acb, bac, bca, cab, cba\}$, $f_{E}(w) = abc$ and $f_{F}(w) = cba$, which crucially means that these two tasks cannot be learnt through any static input-target alignments. Instead, an easier and viable way of learning is through counting the occurrences of each symbol $\sigma \in \Sigma$ for the input sequence in the encoder and then retrieving each $\sigma$, ascendingly or descendingly, according to the previously encoded counts to generate the output sequence in the decoder. The results are shown in Table~\ref{tab:main_res_full_seq_accu_sorting}, which, reaffirm the limited out-of-distribution generalization abilities of RNN seq2seq models, despite their nearly perfect in-distribution performance. Moreover, attention is significantly beneficial for SRNN seq2seq models, but less so for GRU and LSTM models, probably because GRU and LSTM can learn the two sorting tasks through counting even without attention, which SRNN cannot. These explanations are consistent with our previous conclusions.

\begin{table}[ht]
\centering
\small
  {\caption{\label{tab:main_res_full_seq_accu_sorting}Aggregate full-sequence accuracy (\%) for the two sorting tasks.}}
  \vspace{10pt}
  {\begin{tabular}{llrrrrrrrr}
  \toprule
  \multicolumn{2}{c}{} & & \multicolumn{3}{c}{\bfseries Attentional} & & \multicolumn{3}{c}{\bfseries Attention-less} \\ \hline
\bfseries Task & \bfseries Dataset & & \bfseries SRNN & \bfseries GRU & \bfseries LSTM & & \bfseries SRNN & \bfseries GRU & \bfseries LSTM \\ \hline

& Train & & 100.00 & 100.00 & 100.00 && 37.28 & 100.00 & 100.00 \\ 
Ascend & Test & & 99.03 & 99.69 & \bfseries 99.73 && 6.48 & 99.50 & \bfseries 99.74 \\ 
& Gen & & 10.89 & 31.06 & \bfseries 31.43 && 0.02 & \bfseries 42.72 & 35.66 \\ \hline

& Train & & 100.00 & 100.00 & 100.00 && 24.01 & 100.00 & 100.00 \\ 
Descend & Test & & 99.05 & \bfseries 99.78 & 99.69 && 0.49 & 99.19 & \bfseries 99.66 \\ 
& Gen & & 14.65 & 31.12 & \bfseries 32.35 && 0.00 & 34.33 & \bfseries 37.08 \\ 

  \bottomrule
  \end{tabular}}
\end{table}

The results presented above showcase the fruitfulness of utilizing formal language tasks to probe the learning capabilities of neural networks in the context of string transductions. A few key insights or lessons, which are more or less commonsense but worth emphasizing are as follows. (1) Conclusions about learnability should be carefully made, since many factors play a role in the final performance of trained neural networks, such as sample complexity, model complexity, training conditions, etc. For instance, in our main experiments, attention-less SRNN models barely learnt total reduplication, in line with \cite{nelson-etal-2020-probing}, but the follow-up experiment in \cref{sec:attn_learn_res} shows that this is a result of insufficient training resources. (2) Reliably evaluating the empirical learning capabilities of neural networks requires well-rounded evaluation metrics. Had this study only looked at full-sequence accuracy or first $n$-symbol accuracy, it would become unlikely to discover that attentional SRNN seq2seq models learnt a somehow periodic repetition function when trained for quadratic copying. (3) Task complexity is strongly tied to the structure of the learner, as demonstrated by the relative complexity between identity and reversal for FSTs and attention-less RNN seq2seq models. To make this point stronger, we show in Appendix \hyperref[app:rnns_learn_identity]{D.3} that RNNs of few hundred parameters can seemingly learn identity and easily generalize to input sequences of 10,000 symbols, while RNN seq2seq models of millions of parameters cannot. Therefore, we discourage over-interpretations of our results beyond the context of this study.

Besides some unexplained puzzles brought up here, good continuations of the current research may include experimenting with (1) other types of seq2seq models, such as other configuration of RNN seq2seq (e.g., bidirectional encoder), CNN seq2seq \citep{pmlr-v70-gehring17a} and transformer \citep{tranformer}; (2) Tape-RNN, which show promising generalization results in various transduction tasks \citep{nn_chomsky}; (3) and other novel transduction tasks that specifically test the predictions of the complexity hierarchy discussed in 3.3. We hope that our study  encourages more works at the intersection of neural networks and formal language theory in the future.

\acks{The current research would not be initiated and successfully continued without the guidance and inspirations from Prof. Jeffrey Heinz. I am deeply grateful to the three anonymous reviewers from ICGI 2023 for their constructive comments. I also thank Prof. Jordan Kodner, Prof. Thomas Graf, William Oliver, Sarah Payne, Nicholas Behrje who read through the early draft and provided helpful feedback. Parts of the work have been presented at various occasions at Stony Brook University, Yale University, University of Pennsylvania, and George Mason University as a talk or poster in 2023, so my thanks also go for the audiences there.}


\bibliography{references}


\appendix
\section*{A. Mathematical details of the three RNN variants}
\label{app:rnnfs}

For SRNN, $f$ is typically a sigmoid or hyperbolic tangent function. We used the latter. 

\begin{equation}\label{eq:rrn_hidden2}
\mathbf{h}_t = tanh(\mathbf{W}[\mathbf{h}_{t-1}; \mathbf{e}_{t}] + \mathbf{b})
\end{equation}

\noindent 
where $[\; .\; ; \; . \;]$ denotes vector concatenation along the last axis. Since the codomain of $tanh = (-1, 1)$, its exact output may be seen as a signal with a relevancy score $\in [0, 1)$ and a polarity ($\pm$) to compute the next output or the next hidden state. 

For LSTM, $f$ takes what is known as a cell state $\mathbf{c}_{t}$ in addition to the hidden state $\mathbf{h}_{t}$. At each time $t$, both states are updated as follows.

\begin{subequations}\label{eq:lstm_hidden}
\begin{align}
\mathbf{h}_t & = o_{t} * tanh(\mathbf{c}_t) \label{eq:lstm_subeq1} \\
\mathbf{c}_t & = f_{t} * \mathbf{c}_{t-1} + i_{t} * \tilde{\mathbf{c}}_t \label{eq:lstm_subeq2} \\
\tilde{\mathbf{c}}_t & = tanh(\mathbf{W}_c[\mathbf{h}_{t-1}; \mathbf{e}_{t}] + \mathbf{b}_c) \label{eq:lstm_subeq3} \\
o_{t} & = sigmoid(\mathbf{W}_o[\mathbf{h}_{t-1}; \mathbf{e}_{t}] + \mathbf{b}_o) \label{eq:lstm_subeq4} \\
f_{t} & = sigmoid(\mathbf{W}_f[\mathbf{h}_{t-1}; \mathbf{e}_{t}] + \mathbf{b}_f) \label{eq:lstm_subeq5} \\
i_{t} & = sigmoid(\mathbf{W}_i[\mathbf{h}_{t-1}; \mathbf{e}_{t}] + \mathbf{b}_i) \label{eq:lstm_subeq6}
\end{align}
\end{subequations}

\noindent
where Eq.(\ref{eq:lstm_subeq4}-\ref{eq:lstm_subeq6}) are the equations for the output gate $o_{t}$, forget gate $f_{t}$, and input gate $i_{t}$, respectively, all of which are scalar values $\in (0,1)$, thanks to $sigmoid$. Because of this numerical property, they are said to act as a filter to control the information flow inside the recurrent unit. The update of $\mathbf{h}_t$ is determined by $o_{t}$ and $\mathbf{c}_t$, whose update is determined by weighing $\mathbf{c}_{t-1}$ and the temporarily updated cell state $\tilde{\mathbf{c}}_t$ with $f_{t}$ and $i_{t}$, respectively.

Inspired by LSTM, GRU takes a simplified state transition function as follows.

\begin{subequations}\label{eq:gru_hidden}
\begin{align}
\mathbf{h}_t & = z_{t} * \mathbf{h}_{t-1} + (1 - z_{t}) * \tilde{\mathbf{h}}_t \label{eq:gru_subeq1} \\
\tilde{\mathbf{h}}_t & = tanh(\mathbf{W}_h[r_t * \mathbf{h}_{t-1}; \mathbf{e}_{t}] + \mathbf{b}_h) \label{eq:gru_subeq2} \\
z_t & = sigmoid(\mathbf{W}_z[\mathbf{h}_{t-1}; \mathbf{e}_{t}] + \mathbf{b}_z) \label{eq:gru_subeq3} \\
r_{t} & = sigmoid(\mathbf{W}_r[\mathbf{h}_{t-1}; \mathbf{e}_{t}] + \mathbf{b}_r) \label{eq:gru_subeq4}
\end{align}
\end{subequations}

\noindent
where $z_t$ is the update gate, $r_t$ the reset gate, and $\tilde{\mathbf{h}}_t$ the temporarily updated $\mathbf{h}_t$. Here, $\mathbf{h}_t$ takes the place of both $\mathbf{h}_t$ and $\mathbf{c}_t$ as in LSTM, and $z_t$ replaces LSTM's $f_t$ and $i_t$. $\mathbf{h}_t$ is updated by weighing $\mathbf{h}_{t-1}$ and $\tilde{\mathbf{h}}_t$ with $z_t$ and $(1-z_t)$, respectively. 

\begin{table*}[ht]
\centering
\small
  {\caption{\label{tab:data}Summary of data size for the four learning tasks in the main experiments.}}
  \vspace{10pt}
  {\begin{tabular}{cccc}
  \toprule
  \bfseries Dataset & \bfseries Input length & \bfseries \# of pairs per length & \bfseries \# of pairs \\ \hline
   Train & 6-15 & 1,000 & 10,000 \\
   Dev & 6-15 & 1,000 & 10,000 \\ 
   Test & 6-15 & 5,000 & 50,000 \\ 
   Gen & 1-5 \& 16-30 & 5,000 & 100,000 \\
  \bottomrule
  \end{tabular}}
\end{table*}

\section*{B. Experimental details}

\subsection*{B.1. Data}
\label{app:data}

Table~\ref{tab:data} summarizes the data size for the four learning tasks in the main experiments. As mentioned in \cref{sec:data}, duplicates are only allowed for sequences of lengths 1 and 2 in the gen sets to make the later evaluations unified and simpler. For strings of length $l$ (1 or 2), they were randomly sampled from a list of strings $\Sigma^{l} \times m$, where $m$ is the smallest multiplier to repeat $\Sigma^{l}$ $m$ times into a list such that the size of the list $ > 5000$. For strings of longer length $l^{+}$, we uniformly sampled $l^{+}$ symbols from $\Sigma$ with replacement to create a string and repeated this process until we had enough unique strings of length $l^{+}$.  

We reused the original train/dev/test/gen sets for all the follow-up experiments and evaluations with two exceptions. (1) For the follow-up experiment in \cref{sec:attn_learn_res}, the attentional models were trained on 1/4 of data from the original train/dev sets and evaluated on the original test/gen sets. The attention-less models were trained on the train/dev sets that are three times larger than the original ones, with the four datasets re-created from scratch following the same data generation procedure to ensure each dataset remains disjoint from one another. (2) For the follow-up evaluation on the mapping $w \rightarrow w^{40}$ in \cref{sec:rnn_variants} as well as the follow-up experiments in sorting in \cref{sec: discussion}, all the input sequences were directly from the corresponding original datasets. The target sequences were re-created accordingly.

\subsection*{B.2. Model parameter configuration and size}
\label{app:model_size}

Table~\ref{tab:model_size} summarizes the parameter configuration and size of models for all the experiments.

\begin{table*}[ht]
\centering
\footnotesize
  {\caption{\label{tab:model_size}Summary of model parameter configuration and size. H/E: hidden size or embedding size. NA is for models not trained. In the ``Where" row, \cref{sec:results} refers to the main experiments with the rest denoting the respective follow-up experiments.}}
  \vspace{10pt}
  {\begin{tabular}{lrrrrrrrrrr}
  \toprule
  & & \multicolumn{6}{c}{\bfseries Attentional} & & \multicolumn{2}{c}{\bfseries Attention-less} \\ \hline

Where & & \cref{sec:rnn_variants} & \cref{sec:results} \& \cref{sec: discussion} & \cref{sec:attn_learn_res} & \cref{sec:task_complexity} & \cref{sec:task_complexity} & \cref{sec:task_complexity} & & \cref{sec:rnn_variants} & \cref{sec:results} \& \cref{sec: discussion} \\ 

H/E & & 640/384 & 512/128 & 128/128 & 64/128 & 32/128 & 16/128 & & 1024/384 & 512/128 \\ \hline

SRNN & & 2,582,812 & 1,466,396 & 126,236 & 46,236 & 21,596 & 13,116 & & 5,037,084 & 1,204,252 \\ 

GRU  & & NA & 3,305,500 & 291,100 & 104,092 & 44,380 & 22,972 & & NA & 2,519,068  \\ 

LSTM & & NA & 4,225,052 & 373,532 & 133,020 & 55,772 & 27,900 & & NA & 3,176,476 \\ 
  \bottomrule
  \end{tabular}}
\end{table*}

\begin{table*}[ht]
\centering
\small
  {\caption{\label{tab:avg_epochs_run}Average number of epochs used over the three runs for each model configuration across the four tasks in the main experiments.}}
  \vspace{10pt}
  {\begin{tabular}{lrrrrrrrr}
  \toprule
  & & \multicolumn{3}{c}{\bfseries Attentional} & & \multicolumn{3}{c}{\bfseries Attention-less} \\ \hline
\bfseries Task & & \bfseries SRNN & \bfseries GRU & \bfseries LSTM & & \bfseries SRNN & \bfseries GRU & \bfseries LSTM \\ \hline

Identity & & 60 & 93 & 103 & & 500 & 500 & 500 \\ 

Rev  & & 30 & 93 & 90 & &  483 & 500 & 500 \\ 

Total Red & & 83 & 180 & 176 & & 500 & 500 & 500 \\ 

Quad Copy  & & 500 & 500 & 500 & & 500 & 500 & 500 \\ 

Average & & 168 & 216 & 217 & & 496 & 500 & 500 \\ 
  \bottomrule
  \end{tabular}}
\end{table*}

\subsection*{B.3. Training notes}
\label{app:training_notes}

Constrained by computational resources, we were unable to experiment with different sets of hyperparameters, various optimazation methods, and so on. Therefore, we can in no way guarantee that those negative results (e.g., out-of-distribution generalization, SRNN seqseq models can not learn quadratic copying) shown in the paper are not due to our choices of training protocol and model configurations. We only \textbf{assume} that our training and model decisions are reasonably effective and that the obtained results are reasonably representative of the learning capabilities of generic RNN seq2seq models, at least in an empirical sense.  
    
Nevertheless, our training is mostly effective. Table~\ref{tab:avg_epochs_run} provides details about the average number of epochs used over the three runs for each type of model from the main experiments. It shows that when some tasks are easy to learn, particularly for attentional models learning tasks other than quadratic copying, the models converge within 180 epochs on average or less, without unnecessarily running through all epochs. When the tasks are difficult to learn, even when the train set can be overfitted with 100\% full-sequence accuracy, the selection of the best model based on the dev set performance serves as a grid search for the best model out of 50 evaluated models across 500 epochs. With three runs, we try to ensure that the final obtained models are indicative of their respective learning potentials for each task. Finally, the extensive follow-up experiments also help reduce the risks of us being misguided by insufficient training conditions, besides making more fine-grained observations.

\begin{table*}[ht]
\centering
\scriptsize
  {\caption{\label{tab:main_res_full_seq_accu_avg}Average full-sequence accuracy (\%) with standard deviation on the aggregate level across the four learning tasks for models with various configurations. Best average results are in \textbf{bold} for the test and gen sets.}}
  \vspace{10pt}
  {\begin{tabular}{llrrrrrrrr}
  \toprule
  \multicolumn{2}{c}{} & & \multicolumn{3}{c}{\bfseries Attentional} & & \multicolumn{3}{c}{\bfseries Attention-less} \\ \hline
\bfseries Task & \bfseries Dataset & & \bfseries SRNN & \bfseries GRU & \bfseries LSTM & & \bfseries SRNN & \bfseries GRU & \bfseries LSTM \\ \hline

& Train & & 100.00$_{\pm0.01}$ & 100.00$_{\pm0.00}$ & 100.00$_{\pm0.00}$ && 33.05$_{\pm31.81}$ & 98.22$_{\pm0.21}$ & 99.99$_{\pm0.02}$ \\ 
Identity & Test & & 99.88$_{\pm0.12}$ & \bfseries 100.00$_{\pm0.00}$ & 99.99$_{\pm0.01}$ && 15.09$_{\pm24.05}$ & 70.62$_{\pm1.49}$ & \bfseries 76.97$_{\pm0.86}$ \\ 
& Gen & & 18.76$_{\pm5.99}$ & 33.75$_{\pm5.69}$ & \bfseries 33.92$_{\pm2.14}$ && 0.00$_{\pm0.00}$ & 9.67$_{\pm1.20}$ & \bfseries 10.02$_{\pm0.06}$ \\ \hline

& Train & & 100.00$_{\pm0.00}$ & 100.00$_{\pm0.00}$ & 100.00$_{\pm0.00}$ && 100.00$_{\pm0.00}$ & 100.00$_{\pm0.00}$ & 100.00$_{\pm0.00}$ \\ 
Rev & Test & & \bfseries 99.98$_{\pm0.00}$ & 99.93$_{\pm0.05}$ & 99.90$_{\pm0.04}$ && \bfseries 99.57$_{\pm0.02}$ & 88.32$_{\pm0.45}$ & 92.66$_{\pm0.22}$ \\ 
& Gen & & \bfseries 37.51$_{\pm2.97}$ & 21.73$_{\pm1.56}$ & 24.10$_{\pm1.57}$ && \bfseries 19.92$_{\pm4.61}$ & 19.04$_{\pm0.66}$ & 12.44$_{\pm0.21}$ \\ \hline

& Train & & 100.00$_{\pm0.01}$ & 100.00$_{\pm0.01}$ & 100.00$_{\pm0.01}$ && 12.26$_{\pm2.96}$ & 86.99$_{\pm3.32}$ & 92.14$_{\pm1.22}$ \\ 
Total Red & Test & & 99.79$_{\pm0.07}$ & \bfseries 99.84$_{\pm0.07}$ & 99.77$_{\pm0.12}$ && 4.25$_{\pm1.20}$ & 48.86$_{\pm1.65}$ & \bfseries 54.50$_{\pm0.59}$ \\ 
& Gen & & \bfseries 34.98$_{\pm10.03}$ & 19.21$_{\pm4.18}$ & 19.17$_{\pm1.64}$ && 0.00$_{\pm0.00}$ & 3.77$_{\pm0.76}$ & \bfseries 4.81$_{\pm1.27}$ \\ \hline

& Train & & 1.73$_{\pm0.62}$ & 71.12$_{\pm8.15}$ & 66.11$_{\pm14.99}$ && 1.01$_{\pm0.53}$ & 44.91$_{\pm5.45}$ & 56.86$_{\pm9.34}$ \\ 
Quad Copy & Test & & 1.52$_{\pm0.43}$ & 60.25$_{\pm6.87}$ & \bfseries 61.04$_{\pm11.18}$ && 0.37$_{\pm0.21}$ & 24.12$_{\pm3.19}$ & \bfseries 34.21$_{\pm3.82}$ \\ 
& Gen & & 1.62$_{\pm0.23}$ & 5.20$_{\pm2.61}$ & \bfseries 6.41$_{\pm0.59}$ && 0.00$_{\pm0.00}$ & \bfseries 0.74$_{\pm0.39}$ & 0.13$_{\pm0.11}$ \\ \hline

& Train & & 75.43$_{\pm0.16}$ & 92.78$_{\pm2.04}$ & 91.53$_{\pm3.75}$ && 36.58$_{\pm8.82}$ & 82.53$_{\pm2.24}$ & 87.25$_{\pm2.64}$ \\ 
Average & Test & & 75.29$_{\pm0.16}$ & 90.00$_{\pm1.75}$ & \bfseries 90.17$_{\pm2.84}$ && 29.82$_{\pm6.37}$ & 57.98$_{\pm1.69}$ & \bfseries 64.59$_{\pm1.37}$ \\ 
& Gen & & \bfseries 23.22$_{\pm4.80}$ & 19.97$_{\pm3.51}$ & 20.90$_{\pm1.49}$ && 4.98$_{\pm1.15}$ & \bfseries 8.30$_{\pm0.75}$ & 6.85$_{\pm0.41}$ \\
  \bottomrule
  \end{tabular}}
\end{table*}

\section*{C. Results: main experiments}
\label{app:results}

\subsection*{C.1. Average results: full-sequence accuracy}
\label{app:avg_full_seq_accu}

Table~\ref{tab:main_res_full_seq_accu_avg} provides the average results in full-sequence accuracy with standard deviation based on the three runs. Clearly, except few trivial differences, such as which type of model has the best average gen set performance for identity and quadratic copying, Table~\ref{tab:main_res_full_seq_accu_avg} is in line with Table~\ref{tab:main_res_full_seq_accu}. The observations we made from Table~\ref{tab:main_res_full_seq_accu} remain valid in terms of Table~\ref{tab:main_res_full_seq_accu_avg}. While Table~\ref{tab:main_res_full_seq_accu_avg} additionally shows how model performance varies across different runs, the interpretation of the variation may not be meaningful, since such variation also easily ``varies" with changes in hyperparameters. As mentioned in the beginning of \cref{sec:results}, this study is only interested in the learning potentials of RNN seq2seq models. 

The average results in the other two metrics, i.e., first $n$-symbol accuracy and overlap rate, show similar data patterns as the corresponding best results, which are reported in the following two sections. To view the complete sets of results obtained in the paper, raw or summarized, please check the GitHub repository for the study at \url{https://github.com/jaaack-wang/rnn-seq2seq-learning}.

\subsection*{C.2. Best aggregate results measured in other metrics}
\label{app:other_main_results}

Table~\ref{tab:main_res_first_n_accu} and Table~\ref{tab:main_res_overlap_rate} show the aggregate best results measured in first $n$-symbol accuracy and overlap rate, respectively. The results were selected from the same runs consistent with Table~\ref{tab:main_res_full_seq_accu}. Best results are in \textbf{bold} for the test and gen sets.

\begin{table*}[ht]
\centering
\small
  {\caption{\label{tab:main_res_first_n_accu}Aggregate first $n$-symbol accuracy (\%) across the four learning tasks for models with various configurations.}}
  \vspace{10pt}
  {\begin{tabular}{llrrrrrrrr}
  \toprule
  \multicolumn{2}{c}{} & & \multicolumn{3}{c}{\bfseries Attentional} & & \multicolumn{3}{c}{\bfseries Attention-less} \\ \hline
\bfseries Task & \bfseries Dataset & & \bfseries SRNN & \bfseries GRU & \bfseries LSTM & & \bfseries SRNN & \bfseries GRU & \bfseries LSTM \\ \hline

& Train & & 100.00 & 100.00 & 100.00 & & 79.49 & 99.05 & 99.98 \\ 
Identity & Test & & 99.99 & \bfseries 100.00 & \bfseries 100.00 & & 59.55 & 85.64 & \bfseries 90.58 \\ 
& Gen & & 65.51 & \textbf{74.83} & 71.89 & & 6.90 & 43.44 & \textbf{50.19} \\ \hline

& Train & & 100.00 & 100.00 & 100.00 & & 100.00 & 100.00 & 100.00 \\ 
Rev & Test & & \bfseries 99.99 & 99.93 & 99.93 & & \bfseries 99.94 & 96.26 & 97.66 \\ 
& Gen & & \textbf{64.01} & 35.52 & 36.88 & & \textbf{69.69} & 45.58 & 53.39 \\ \hline

& Train & & 100.00 & 100.00 & 100.00 & & 36.66 & 93.99 & 96.09 \\ 
Total Red & Test & & 99.89 & \bfseries 99.90 & 99.83 & & 24.27 & 66.85 & \bfseries 70.68 \\ 
& Gen & & \textbf{73.43} & 50.15 & 50.40 & & 2.52 & 20.60 & \textbf{21.44} \\ \hline

& Train & & 87.67 & 99.70 & 98.73 && 17.86 & 83.48 & 92.65 \\ 
Quad Copy  & Test & & 85.39 & \bfseries 98.49 & 97.01 && 10.94 & 49.85 & \bfseries 60.52 \\ 
& Gen & & \bfseries 76.42 & 59.33 & 29.68 && 0.85 & \bfseries 12.43 & 9.86 \\ \hline

& Train & & 96.92 & 99.93 & 99.68 && 58.50 & 94.13 & 97.18 \\ 
Average & Test & & 96.32 & \bfseries 99.58 & 99.19 && 48.68 & 74.65 & \bfseries 79.86 \\ 
& Gen & & \bfseries 69.84 & 54.96 & 47.21 && 19.99 & 30.51 & \bfseries 33.72 \\ 
  \bottomrule
  \end{tabular}}
\end{table*}

\begin{table*}[ht]
\centering
\small
  {\caption{\label{tab:main_res_overlap_rate}Aggregate overlap rate (\%) across the four learning tasks for models with various configurations.}}
  \vspace{10pt}
  {\begin{tabular}{llrrrrrrrr}
  \toprule
  \multicolumn{2}{c}{} & & \multicolumn{3}{c}{\bfseries Attentional} & & \multicolumn{3}{c}{\bfseries Attention-less} \\ \hline
\bfseries Task & \bfseries Dataset & & \bfseries SRNN & \bfseries GRU & \bfseries LSTM & & \bfseries SRNN & \bfseries GRU & \bfseries LSTM \\ \hline

& Train & & 100.00 & 100.00 & 100.00 & & 89.31 & 99.63 & 99.99 \\ 
Identity & Test & & \bfseries 100.00 & \bfseries 100.00 & \bfseries 100.00 & & 76.51 & 92.71 & \bfseries 95.56 \\ 
& Gen & & 68.61 & \textbf{80.31} & 78.09 & & 12.34 & 50.99 & \textbf{58.55} \\ \hline

& Train & & 100.00 & 100.00 & 100.00 & & 100.00 & 100.00 & 100.00 \\ 
Rev & Test & & \bfseries 100.00 & 99.97 & 99.97 & & \bfseries 99.97 & 98.02 & 98.92 \\ 
& Gen & & \textbf{69.36} & 39.65 & 41.95 & & \textbf{72.77} & 53.57 & 61.99\\ \hline

& Train & & 100.00 & 100.00 & 100.00 & & 59.39 & 97.80 & 98.74 \\ 
Total Red & Test & & \bfseries 99.96 & \bfseries 99.96 & 99.95 & & 47.89 & 83.30 & \bfseries 85.98  \\ 
& Gen & & \textbf{83.09} & 58.00 & 54.39 & & 8.49 & 30.96 & \textbf{32.28} \\ \hline

& Train & & 98.45 & 99.71 & 98.85 && 29.82 & 89.65 & 95.76 \\ 
Quad Copy  & Test & & 98.22 & \bfseries 98.60 & 97.91 && 23.24 & 64.59 & \bfseries 76.24 \\ 
& Gen & & \bfseries 88.02 & 64.30 & 40.64 && 6.16 & \bfseries 19.53 & 16.54 \\ \hline

& Train & & 99.61 & 99.93 & 99.71 && 69.63 & 96.77 & 98.62 \\ 
Average & Test & & 99.54 & \bfseries 99.63 & 99.46 && 61.90 & 84.65 & \bfseries 89.17 \\ 
& Gen & & \bfseries 77.27 & 60.56 & 53.77 && 24.94 & 38.76 & \bfseries 42.34 \\ 
  \bottomrule
  \end{tabular}}
\end{table*}

\subsection*{C.3. Best per-input-length results measured in other metrics}
\label{app:other_per_seq_perf}

\begin{figure}[!htb]
\centering
  \includegraphics[width=1\columnwidth]{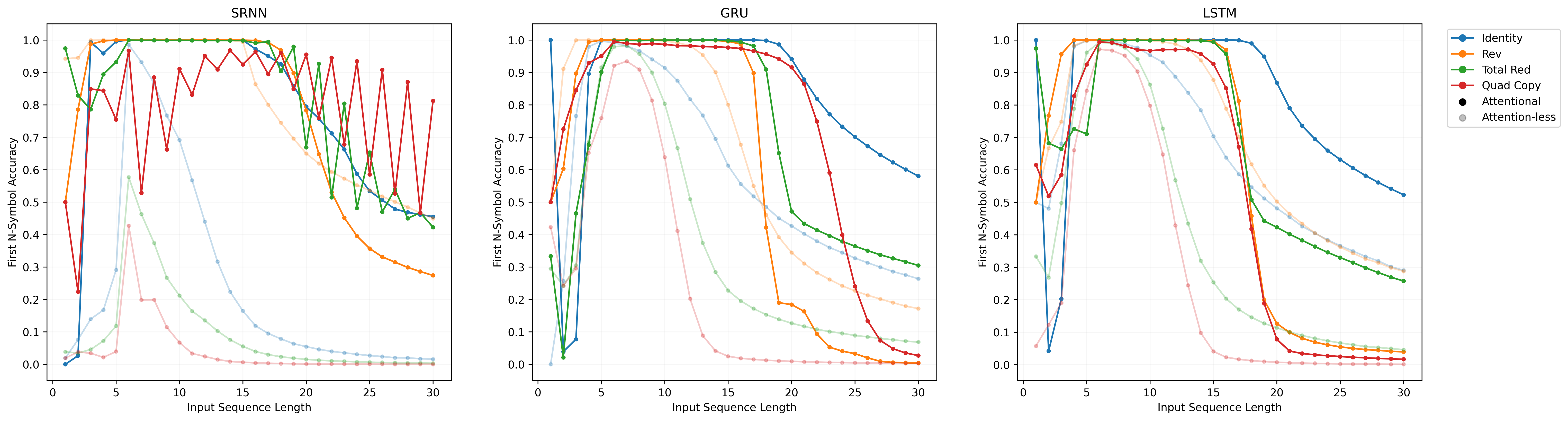}
\caption{Test/gen set first $n$-symbol accuracy per input length across the four tasks for the three types of RNN seq2seq models, with and without attention.}
\label{fig:first_n_accu_per_len}
\end{figure}

\begin{figure}[!htb]
\centering
  \includegraphics[width=1\columnwidth]{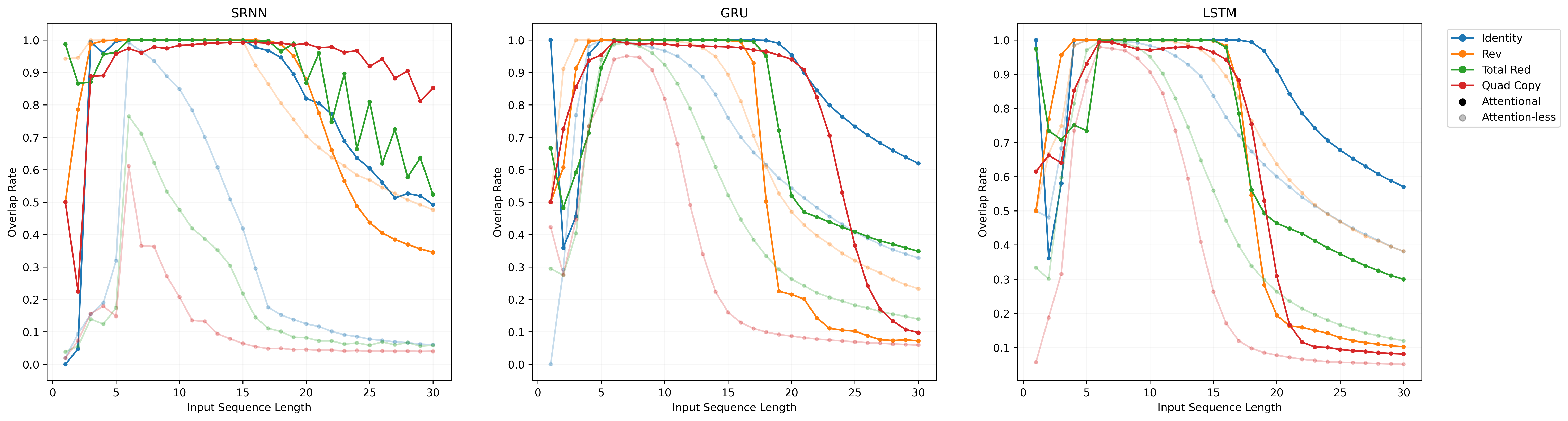}
\caption{Test/gen set overlap rate per input length across the four tasks for the three types of RNN seq2seq models, with and without attention.}
\label{fig:overlap_rate_per_len}
\end{figure}

Fig~\ref{fig:first_n_accu_per_len} and Fig~\ref{fig:overlap_rate_per_len} show the main results measured in first $n$-symbol accuracy and overlap rate on the per-input-length level, respectively. Please note that, the reason why the best attentional SRNN model outperforms the related GRU and LSTM counterparts in terms of the overall test/gen set performance is not because it learns quadratic copying. Instead, attentional SRNN models all ended up learning a function that more or less repeats the input sequences without knowing when exactly to produce the end symbol $<\!s\!>$. Please check \cref{sec:rnn_variants} for the reasoning.

\section*{D. Results: follow-up experiments}
\label{app:results_follow_up}

\subsection*{D.1. Overview}

Follow-up experiments with RNN seq2seq models were conducted in the following sections: (1) \cref{sec:attn_learn_res}, where we contrasted attentional models with attention-less ones for their efficiency in learning total reduplication; (2) \cref{sec:task_complexity}, where we further examined the relative complexity among identity, reversal, and total reduplication for attentional models; (3) \cref{sec:rnn_variants}, where we studied the possibility of SRNN seq2seq models, with or without attention, learning quadratic copying; (4) and \cref{sec: discussion}, where we experimented with ascending and descending sorting, to test the generality of the findings based on the four tasks in the main experiments. 

These follow-up experiments all follow the same training and evaluation procedures as the main experiments. That is to say, for each experiment, we trained and evaluated each model for three runs and selected the best results for interpretation. Details about data can be found in Appendix \hyperref[app:data]{B.1}. Details about model parameter configuration and size are in Appendix \hyperref[app:model_size]{B.2}. For more details, please check the GitHub repository for the paper. 

In what follows, we provide some further details about the follow-up experiments in \cref{sec:rnn_variants}, mentioned above. We also introduce another follow-up experiment with RNNs trained for identity mentioned in \cref{sec: discussion}, but not above. Both are for the self-containedness of the paper.

\subsection*{D.2. Attentional SRNN seq2seq models trained for quadratic copying \label{app:attn_srnn_quad_copy}}

Table~\ref{tab:larger_srnn_quad_copy} shows the performance of the best enlarged SRNN models trained and evaluated on the original datasets created for quadratic copying. Clearly, neither fit the train set and the one without attention is not even learning anything because its overlap rate is just about random guess, which equals $\frac{1}{|\Sigma|} = \frac{1}{26} = 3.85\%$, or $\frac{1}{28} = 3.57\%$ plus the two edge symbols. This is despite the attention-less SRNN model having more than 5 million parameters, the largest model trained in the study. We intended to make the attentional SRNN model larger than the related GRU or LSTM model, but it was impossible without modifying the model structure. Increasing hidden size too much exceeded the maximum 40G GPU RAM on Google Colaboratory due to too much computation for softmax in the attention layer.

\begin{table*}[ht]
\centering
\small
  {\caption{\label{tab:larger_srnn_quad_copy}The performance across the three metrics (\%) for the best enlarged SRNN models trained for quadratic copying.}}
  \vspace{10pt}
  {\begin{tabular}{lcccccccc}
  \toprule
  & \multicolumn{3}{c}{\bfseries Attentional} & & \multicolumn{3}{c}{\bfseries Attention-less} \\ \hline
  
  \bfseries Dataset & \bfseries Full-seq & \bfseries First $n$-symbol & \bfseries Overlap & & \bfseries Full-seq & \bfseries First $n$-symbol & \bfseries Overlap \\ \hline
  Train & 3.43 & 92.43 & 98.65 & & 0.00 & 0.05 & 3.80 \\ 
   Test & 3.00 & 90.92 & 98.53 & & 0.00 & 0.05 & 3.81 \\ 
   Gen & 2.79 & 84.23 & 92.82 & & 0.00 & 0.19 & 3.68 \\
  \bottomrule
  \end{tabular}}
\end{table*}

\begin{figure}[!htb]
\centering
  \includegraphics[width=1\columnwidth]{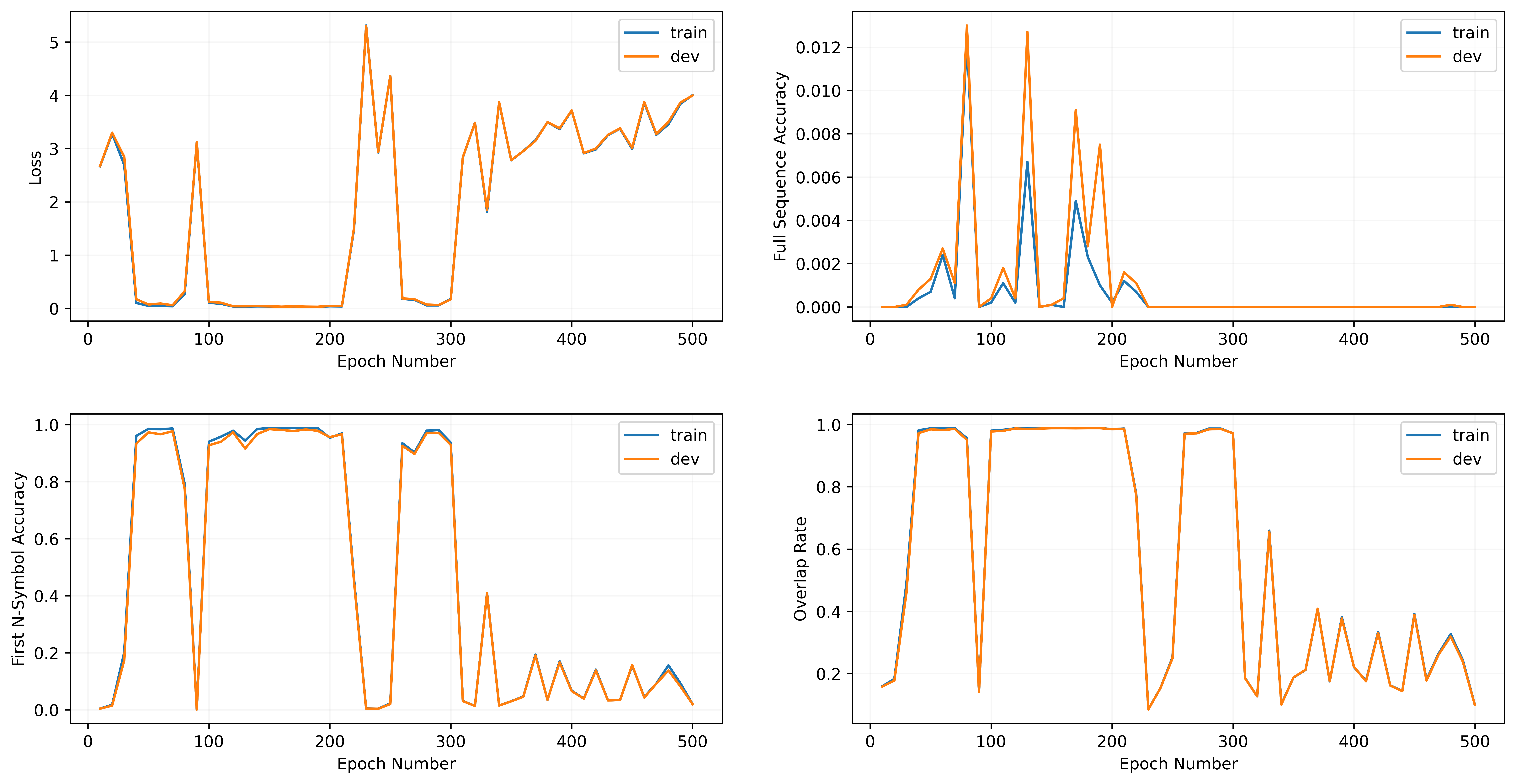}
\caption{Training log of the third attentional SRNN model for quadratic copying in the main experiment, i.e., the ``worst" SRNN model in Table~\ref{tab:40_copies_first_n}.}
\label{fig:training_plot_attn_srnn_quad_copy}
\end{figure}

In line with Tables~\ref{tab:main_res_first_n_accu} and~\ref{tab:main_res_overlap_rate}, Table~\ref{tab:larger_srnn_quad_copy} also shows unusually high first $n$-symbol accuracy and overlap rate for attentional SRNN models trained for quadratic copying. Moreover, we found that during training, all the six attentional SRNN models, including the three enlarged ones, could easily reach nearly 100.00\% first $n$-symbol accuracy (and thus overlap rate) within 100 epochs and that first $n$-symbol accuracy is not correlated with full-sequence accuracy, unlike all the rest models we trained that all show a rough correlation. Fig~\ref{fig:training_plot_attn_srnn_quad_copy} is an example training log of an non-enlarged attentional SRNN model for quadratic copying. 

The poor correlation is probably because attentional SRNN models are approximating a somehow periodic function that repeats the input sequence for either a fixed or unbounded number of times, which conflicts with learning quadratic copying, which requires a bounded and yet varied number of repetitions of the input sequence according to the input length. The fact that the trained attentional SRNN models tend to have a unusually high first $n$-symbol accuracy means that the approximated periodic function, if perfectly learnt, should ideally repeat any given input sequence more than 30 times, i.e., the maximum input length available in the gen set.

Therefore, we tested all the saved attentional models trained for quadratic copying on the mapping $w \rightarrow w^{40}$. The results in Table~\ref{tab:40_copies_first_n} shows that even the ``worst-performing" SRNN model outperforms the ``best-performing" GRU/LSTM model in terms of first $n$-symbol accuracy, given that full-sequence accuracy is always 0.00\%. Since the saved attentional SRNN models were selected based on the best dev set full-sequence accuracy, we expect that all these SRNN models would have shown much better first $n$-symbol accuracy on $w \rightarrow w^{40}$ if the best first $n$-symbol accuracy was used instead. As can be seen in Fig~\ref{fig:training_plot_attn_srnn_quad_copy}, the train-dev variance in first $n$-symbol accuracy is negligible and thus the test set performance should come close. That said, the exact function learnt by SRNN models remains unknown.

\begin{table*}[ht]
\centering
\small
  {\caption{\label{tab:40_copies_first_n}The test/gen set first $n$-symbol accuracy (\%) for all the attentional models trained for quadratic copying across three runs on the mapping $w \rightarrow w^{40}$.}}
  \vspace{10pt}
  {\begin{tabular}{lrrrrrrrr}
  \toprule
  & \multicolumn{3}{c}{\bfseries Test} & & \multicolumn{3}{c}{\bfseries Gen} \\ \hline
  
  \bfseries Model & \bfseries Run\#1 & \bfseries Run\#2 & \bfseries Run\#3 & & \bfseries Run\#1 & \bfseries Run\#2 & \bfseries Run\#3 \\ \hline
  SRNN & 67.95 & 84.16 & 68.33 & & 67.07 & 68.42 & 30.89 \\ 
   SRNN$_{Large}$ & 84.86 & 82.14 & 96.20 & & 62.89 & 71.70 & 80.81 \\ 
   GRU & 26.42 & 25.49 & 26.82 & & 23.66 & 10.67 & 14.15 \\
   LSTM & 26.83 & 25.51 & 25.52 & & 6.07 & 8.72 & 7.56 \\
  \bottomrule
  \end{tabular}}
\end{table*}

\subsection*{D.3. RNNs of few hundred parameters learning identity \label{app:rnns_learn_identity}}

The RNNs we trained here are essentially the decoder part of RNN seq2seq models without attention. The major difference between them is that RNNs work just like FSTs, instead of auto-regressively, so they produce an output symbol immediately after consuming an input symbol, until the entire input sequence is read. 

We trained SRNN, GRU, and LSTM with embedding size 3 and hidden size 4 and the parameter size for these three models are only 244, 316, and 352, respectively. The training conditions were mostly identical to the main experiments, except for the much larger learning rate, i.e., 1e-2. Apart from 150,000 strings of lengths 1-30 from the original test and gen sets, we also evaluated the trained models on 10,000 random unique strings of length 10,000. All the models we trained achieved 100\% full-sequence accuracy on these unseen 160,000 strings. Since RNNs produce as many symbols as they consume, the result also means that these models can transduce all the prefixes of those 10,000 strings, ranging from length 1 to 9,999, successfully. See: \url{https://github.com/jaaack-wang/RNNs-learn-identity}.

The experiment presented above simply serves as a proof of concept. The result is actually not surprising because for RNNs to learn identity, they only need to produce whatever they consume. Although RNNs can learn identity much better than RNN seq2seq models, it is easy to prove that unidirectional RNNs cannot learn at all functions that require 2-way FSTs, because they cannot look ahead.\footnote{Please note that, here we only refer to canonical RNNs that work like FSTs. RNNs can be trained to consume all input symbols first and then use the final hidden state to auto-regressively produce output symbols. However, RNNs trained in this way can be seen as a special case of attention-less RNN seq2seq models where encoder and decoder share the same weights. RNNs trained in \cite{nn_chomsky} are similar to this, but instead of doing auto-regression, they let the models consume as many dummy tokens as the target sequence after they consumed the entire real input sequence, during which the models produce an output symbol once they consume an dummy token, just like a FST. Nevertheless, neither of these two special types of ``RNNs" are considered as RNNs in the sense of this study.} The point here is to show that there is a strong connection between task complexity and the structure of the learner and that RNNs and RNN seq2seq models represent two classes of neural networks (also see \cref{sec:RNN_vs_RNN_seq2seq}). Other than the aforementioned point, we do not make any more claims here.

\end{document}